\documentclass{article}

     \usepackage[preprint]{neurips_2024}

\usepackage[utf8]{inputenc} 
\usepackage[T1]{fontenc}    
\usepackage{hyperref}       
\usepackage{url}            
\usepackage{booktabs}       
\usepackage{amsfonts}       
\usepackage{nicefrac}       
\usepackage{microtype}      
\usepackage{xcolor}         

\usepackage{color}
\usepackage{babel}
\usepackage{array}
\usepackage{amsmath}
\usepackage{amsthm}
\usepackage{amssymb}
\usepackage{graphicx}

\title{EKM: an exact, polynomial-time algorithm for the $K$-medoids problem }

\author{%
  Xi He\\
  School of Computer Science\\
  University of Birmingham\\
  Birmingham, B15 2TT, UK \\
  \texttt{xxh164@student.bham.ac.uk} \\
   \And
   Max A. Little \\
   School of Computer Science\\
   University of Birmingham\\
   Birmingham, B15 2TT, UK\\
   \texttt{maxl@mit.edu} \\
}

\begin{document}

\maketitle

\begin{abstract}
The $K$-medoids problem is a challenging combinatorial clustering
task, widely used in data analysis applications. While numerous algorithms
have been proposed to solve this problem, none of these are able to
obtain an exact (globally optimal) solution for the problem in polynomial
time. In this paper, we present EKM: a novel algorithm for solving
this problem exactly with worst-case $O\left(N^{K+1}\right)$ time
complexity. EKM is developed according to recent advances in transformational
programming and combinatorial generation, using formal program derivation
steps. The derived algorithm is provably correct by construction.
We demonstrate the effectiveness of our algorithm by comparing it
against various approximate methods on numerous real-world datasets.
We show that the wall-clock run time of our algorithm matches the
worst-case time complexity analysis on synthetic datasets, clearly
outperforming the exponential time complexity of benchmark branch-and-bound
based MIP solvers. To our knowledge, this is the first, rigorously-proven
polynomial time, practical algorithm for this ubiquitous problem.
\end{abstract}

\section{Introduction}

In machine learning (ML), $K$-medoids is the problem of partitioning
a given dataset into $K$ clusters where each cluster is represented
by one of its data points, known as a \emph{medoid}. A plethora of
approximate/heuristic methodologies have been proposed to solve this
problem, such as PAM (Partitioning Around Medoids) \citep{kaufman1990partitioning},
CLARANS (Clustering Large Applications based on randomized Search)
\citep{ng2002clarans}, and variants of PAM \citep{schubert2021fast,van2003new}.

While approximate algorithms are computationally efficient and have
been scaled up to large dataset sizes, because they are approximate
they provide no guarantee that the \emph{exact} (\emph{globally optimal})\emph{
	solution} can be obtained. In \emph{high-stakes} or \emph{safety-critical
}applications, where errors are unacceptable or carry significant
costs, we want the best possible partition given the specification
of the clustering problem. Only an exact algorithm can provide this
guarantee.

However, there are relatively few studies on exact algorithms for
the $K$-medoids problem. The use of \emph{branch-and-bound} (BnB)
methods predominates research on this problem \citep{ren2022global,elloumi2010tighter,christofides1982tree,ceselli2005branch}.
An alternative approach is to use off-the-shelf \emph{mixed-integer
	programming solvers }(MIP) such as Gurobi \citep{gurobi2021gurobi}
or GLPK (GNU Linear Programming Kit) \citep{makhorin2008glpk}. These
solvers have made significant achievements, for instance, \citep{elloumi2010tighter,ceselli2005branch}'s
BnB algorithm is capable of processing medium-scale datasets with
a very large number of medoids. More recently, \citep{ren2022global}
designed another BnB algorithm capable of delivering tight approximate
solutions---with an optimal gap of less than 0.1\%---on very large-scale
datasets, comprising over one million data points with three medoids,
although this required a massively parallel computation over 6,000
CPU cores.

These existing studies on the exact $K$-medoids problem have three
defects. Firstly, most previous studies on exact algorithms impose
a computation time limit, thus exact solutions are rarely actually
calculated; the only rigorously exact cases are those computed by
\citep{christofides1982tree,ceselli2005branch} for small datasets
with a maximum size of 150 data items. Secondly, worst-case time and
space complexity analyses are not reported in studies of BnB-based
algorithms. Such BnB methods have exponential worst-case time/space
complexity, yet this critical aspect is rarely discussed or analyzed
rigorously. As a result, existing studies on the exact algorithms
for the $K$-medoids problem either omit time complexity analysis
\citep{elloumi2010tighter,christofides1982tree,ceselli2005branch}
or overlook essential details \citep{ren2022global} such as algorithm
complexity with respect to cluster size and the impact on performance
of upper bound tightness. This omission undermines the reproducibility
of findings and hinders advancement of knowledge in this domain. Thirdly,
proofs of correctness of these algorithms is often omitted. Exact
solutions require such mathematical proof of global optimality, yet
many BnB algorithm studies rely on weak assertions or informal explanations
that do not hold up under close scrutiny \citep{fokkinga1991exercise}.

In this report, we take a fundamentally different approach, by combining
several modern, broadly applicable algorithm design principles from
the theories of \emph{transformational programming} \citep{meertens1986algorithmics,jeuring1993theories,bird1996algebra}
and\emph{ combinatorial generation} \citep{kreher1999combinatorial}.
The derivation of our algorithm is obtained through a rigorous, structured
approach known as\emph{ }the \emph{Bird-Meertens formalism }(or\emph{
}the\emph{ algebra of programming}) \citep{meertens1986algorithmics,jeuring1993theories,bird1996algebra}.
This formalism enables the development of an efficient and correct
algorithm by starting with an obviously correct but possibly inefficient
algorithm and deriving an equivalent, efficient implementation by
way of \emph{equational reasoning} steps. Thus, the correctness of
the algorithm is assured, skipping the need for error-prone, post-hoc
induction proofs. Our proposed algorithm for the $K$-medoids problem
has worst-case $O\left(N^{K+1}\right)$ time complexity and furthermore
is amenable to optimally efficient parallelization\footnote{In parallel computing, an embarrassingly parallel algorithm (also
	called embarrassingly parallelizable, perfectly parallel, delightfully
	parallel or pleasingly parallel) is one that requires no communication
	or dependency between the processes.}.

The paper is organized as follows. In Section 2, we explain in detail
how our efficient EKM algorithm is derived through provably correct
equational reasoning steps. Section 3 shows the results of empirical
computational comparisons against approximate algorithms applied to
data sets from the UCI machine learning repository, and reports time-complexity
analysis in comparison with a MIP solver (GLPK). Section 4 summarizes
the contributions of this study, reviews related work. Finally, Section
5 suggests future research directions.

\section{Theory}

Our novel EKM algorithm for $K$-medoids is derived using rational
algorithm design steps, described in the following subsections. First,
we formally specify the problem as a MIP (Subsection \ref{subsec:spec-MIP}).
Next, we represent an \emph{exhaustive search }algorithm for finding
a globally optimal solution to this MIP in \emph{generate-evaluate-select}
form (Subsection \ref{subsec:exhaustive}). Then, we construct an
efficient, recursive combinatorial configuration generator (Subsection
\ref{subsec:generator}), whose form makes the exhaustive algorithm
amenable to equational program transformation steps which maximize
the implementation efficiency through the use of \emph{shortcut fusion
	theorems }(Subsection \ref{subsec:fusion}).

There are two main advantages to this approach of deriving correct
(globally optimal) algorithms from specifications. First, since the
exhaustive search algorithm is correct for the MIP problem, any algorithm
derived through correct equational reasoning steps from this specification
is also (provably) correct. Second, the proof is short and elegant:
a few generic theorems for shortcut fusion already exist and when
the conditions of these shortcut fusion theorems hold, it is simple
to apply them to a specific problem such as the $K$-medoids problem
here.

\subsection{Specifying a mixed-integer program (MIP) for the $K$-medoids problem\label{subsec:spec-MIP}}

We denote a data set $\mathcal{D}$ consists of $N$ \emph{data points}
$\boldsymbol{x}_{n}=\left(x_{n1},x_{n2},\ldots,x_{nD}\right)\in\mathbb{R}^{D}$,
$\forall n\in\left\{ 1,\ldots,N\right\} =\mathcal{N}$, where $D$
is the dimension of the\emph{ feature space}. We assume the data items
$\boldsymbol{x}_{n}\in\mathcal{D}$ are all stored in a ordered sequence
(list) $\mathcal{D}=\left[\boldsymbol{x}_{1},\boldsymbol{x}_{2},\ldots,\boldsymbol{x}_{N}\right]$.
In clustering problem, we need to find a set of centroids $\mathcal{U}=\left\{ \boldsymbol{\mu}_{k}:k\in\mathcal{K}\right\} $,
where $\mathcal{K}=\left\{ 1,\ldots,K\right\} $ in $\mathbb{R}^{D}$,
and each centroid $\boldsymbol{\mu}_{k}$ associated with a unique
\emph{cluster $C_{k}$}, which subsumes all data points $\boldsymbol{x}$
are the \emph{closest} to this centroid than other centroids, this
closeness is defined by a distance function $d\left(\boldsymbol{x},\boldsymbol{\mu}\right)$,
there is no constrains on the distance function in the $K$-medoids
problem, a common choice is the \emph{squared Euclidean distance function}
$d_{2}\left(\boldsymbol{x},\boldsymbol{\mu}\right)=\left\Vert \boldsymbol{x}-\boldsymbol{y}\right\Vert _{2}^{2}$
. For each set of centroids $\mathcal{U}$ we have a set of \emph{disjoint
	clusters} $\mathcal{C}=\left\{ C_{1},C_{2},\ldots C_{K}\right\} $,\emph{
}and\emph{ }set $\mathcal{K}$ is then called the \emph{cluster labels.}

The $K$-medoids problem is usually specified as the following MIP

\begin{equation}
	\begin{aligned}\mathcal{U}^{*}=\underset{\mathcal{U}}{\textrm{argmin }} & E\left(\mathcal{U}\right)\\
		\text{subject to } & \mathcal{U}\subseteq\mathcal{D},\left|\mathcal{U}\right|=K,
	\end{aligned}
	\label{eq:K-medoids_MIP}
\end{equation}
where $E\left(\mathcal{U}\right)=\sum_{k\in\mathcal{K}}\sum_{x_{n}\in C_{k}}d\left(\boldsymbol{x}_{n},\boldsymbol{\mu}_{k}\right)$
is the \emph{objective function }for the $K$-medoids problem, and
$\mathcal{U}^{*}$ is a set of centroids that optimize the objective
function $E\left(\mathcal{U}\right)$. In our study, we make no assumptions
about the objective function, allowing any arbitrary distance function
to be applied without affecting the worst-case time complexity while
still obtaining the exact solution.

\subsection{Representing the solution as an (inefficient) exhaustive search algorithm\label{subsec:exhaustive}}

In the theory of transformational programming \citep{bird1996algebra,jeuring1993theories},
combinatorial optimization problems such as (\ref{eq:K-medoids_MIP})
are solved using the following, generic, \emph{generate-evaluate-select}
algorithm,

\begin{equation}
	s^{*}=\mathit{sel}_{E}\left(\mathit{eval}_{E}\left(\mathit{gen}\left(\mathcal{D}\right)\right)\right).\label{eq:generate-and-test_paradigm}
\end{equation}

Here, the generator function $\mathit{gen}:\mathit{\mathcal{D}}\to\left[\left[\mathbb{R}^{D}\right]\right]$,
enumerates \emph{all} possible \emph{combinatorial configurations}
$s:\left[\mathbb{R}^{D}\right]$ (here, configurations consist of
a list of data items representing the $K$ medoids) within the \emph{solution
}(\emph{search})\emph{ space}, $\mathcal{S}$ (here, this is the set
of all possible size $K$ combinations). For most problems, $\mathit{gen}$
is a recursive function so that the input of the generator can be
replaced with the index set $\mathcal{N}$ and rewritten $\mathit{gen}\left(n\right)$,
$\forall n\in\mathcal{N}$. The \emph{evaluator} $\mathit{eval}_{E}:\left[\left[\mathbb{R}^{D}\right]\right]\to\left[\left(\left[\mathbb{R}^{D}\right],\mathbb{R}\right)\right]$
computes the objective values $r=E\left(s\right)$ for all configurations
$s$ generated by $\mathit{gen}\left(n\right)$ and returns a list
of \emph{tupled configurations} $\left(s,r\right)$. Lastly, the \emph{selector}
$\mathit{sel}_{E}:\left[\left(\left[\mathbb{R}^{D}\right],\mathbb{R}\right)\right]\to\left(\left[\mathbb{R}^{D}\right],\mathbb{R}\right)$
select the best configuration $s^{*}$ with respect to objective $E$.

Algorithm (\ref{eq:generate-and-test_paradigm}) is an example of
\emph{exhaustive }or \emph{brute-force search}: by generating all
possible configurations in the search space $\mathcal{S}$ for (\ref{eq:K-medoids_MIP}),
evaluating the corresponding objective $E$ for each, and selecting
an optimal configuration, it is clear that it must solve the problem
(\ref{eq:K-medoids_MIP}) exactly. Taking a different perspective,
(\ref{eq:generate-and-test_paradigm}) can be considered as a generic
\emph{program} for solving the MIP (\ref{eq:K-medoids_MIP})\emph{.
}However, program (\ref{eq:generate-and-test_paradigm}) is generally
inefficient due to \emph{combinatorial explosion}; the size of $\mathit{gen}\left(\mathcal{D}\right)$
is often exponential (or worse) in the size of $\mathcal{D}$.

To make this exhaustive solution practical, a generic program transformation
principle known \emph{shortcut fusion} can be used to derive a computationally
tractable program for (\ref{eq:generate-and-test_paradigm}). Efficiency
is achieved by recognizing that in many combinatorial problems, generator
functions can be given as efficient recursions, for instance taking
$O\left(N\right)$ steps. This recursive form makes it possible to
combine or \emph{fuse }the recursive evaluator, and sometimes also
the recursive selector, directly into the recursive generator. This
fusion can save substantial amounts of computation because it eliminates
the need to generate and store every configuration. The evaluated
configurations are generated and selected through a single, fused,
efficient recursive program.

\subsection{Constructing an efficient, recursive combinatorial configuration
	generator\label{subsec:generator}}

In $K$-medoids there are only $N\times\left(N-1\right)\times\cdots\times\left(N-K\right)$
ways of selecting centroids whose corresponding assignments are potentially
distinct. Thus, the solution space $\mathcal{S}$ for the $K$-medoids
problem consists of all size $K$ \emph{combinations} ($K$\emph{-sublists})
of the data input $\mathcal{D}$, denoted as $s:\left[\mathbb{R}^{D}\right]$.
If we can design an efficient, recursive combination generator $\mathit{gen}_{\mathit{combs}}:\mathcal{N}\times\mathcal{K}\to\left[\mathbb{R}^{D}\right]$
to enumerate all possible size $K$ combinations, at least one set
of such centroids corresponds to an optimal value of the clustering
objective $E$. An efficient and structured combination generator
which achieves this, will be described next.

\begin{figure}
	\begin{centering}
		\includegraphics[scale=0.25]{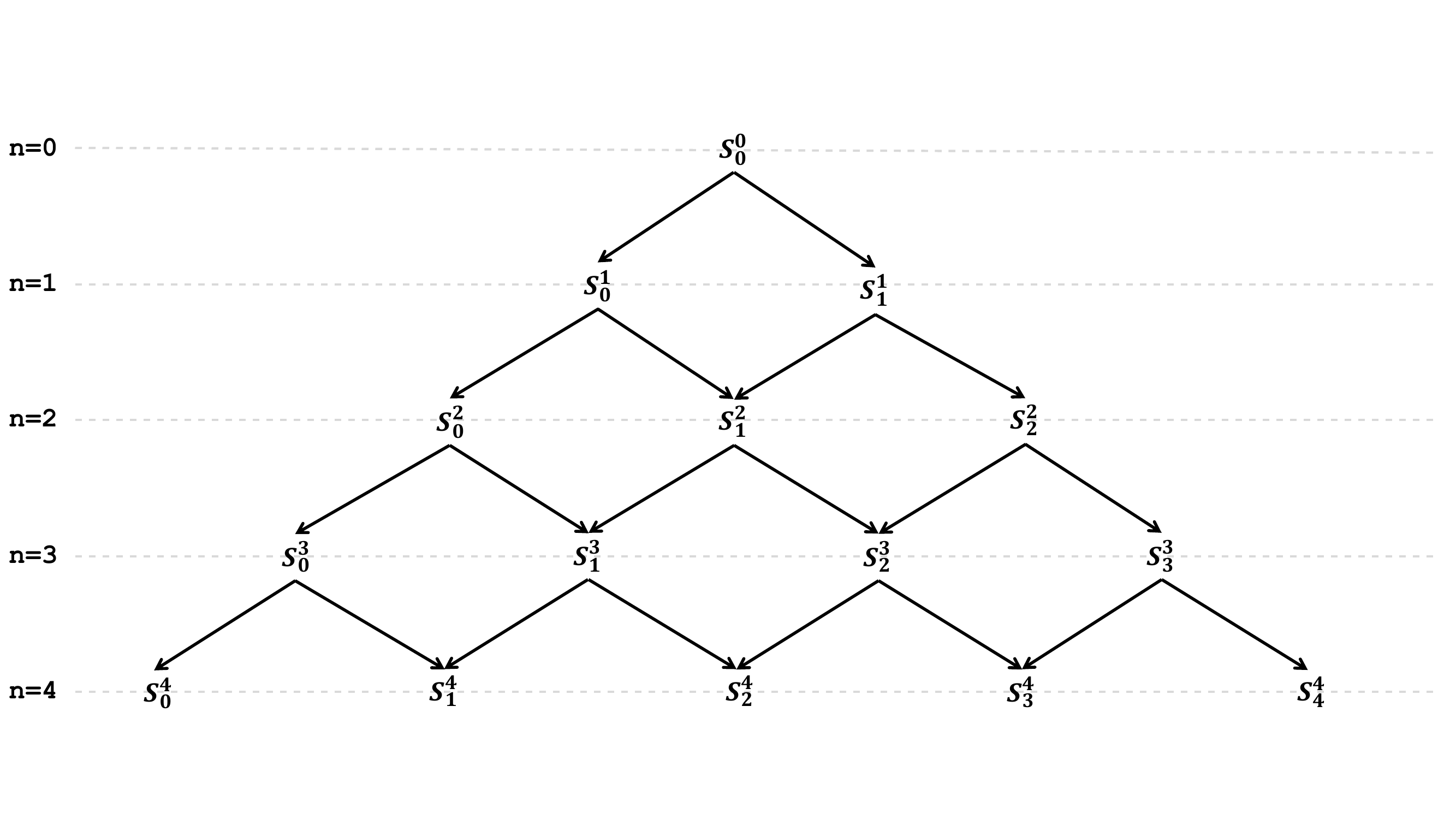}
		\par\end{centering}
	\caption{A generic recursive generator of all possible sublists for list $\left[\boldsymbol{x}_{1},\boldsymbol{x}_{2},\boldsymbol{x}_{3},\boldsymbol{x}_{4}\right]$,
		in each recursive stage $n$, the sublists of size $k$ (combinations)
		are automatically grouped into list $\mathcal{S}_{k}^{n}$, for all
		$k\in\left\{ 0,1,\ldots n\right\} $. \label{fig:generic-recursive-generator }}
\end{figure}

Denoting $\mathcal{S}_{k}^{n}:\left[\mathbb{R}^{D}\right]$ as the
list that stores all possible size $k$ combinations from list $\left[\boldsymbol{x}_{1},\boldsymbol{x}_{2},\ldots\boldsymbol{x}_{n}\right]$
of length $n$, and $\mathcal{S}^{n}=\left[\mathcal{S}_{k}^{n}\mid k\in\left\{ 0,\ldots n\right\} \right]$
as the list of all \emph{sublists} of that list. The key to constructing
an efficient combination generator can be reduced to solving the following
problem: given all sublists $\mathcal{S}^{n}$ for list $\mathcal{D}_{1}=\left[\boldsymbol{x}_{1},\boldsymbol{x}_{2},\ldots\boldsymbol{x}_{n}\right]$,
and $\mathcal{S}^{m}$ for list $\mathcal{D}_{2}=\left[\boldsymbol{y}_{1},\boldsymbol{y}_{2},\ldots\boldsymbol{y}_{m}\right]$,
construct all sublists $\mathcal{S}^{n+m}$ for list $\mathcal{D}_{1}\cup\mathcal{D}_{2}$.
This problem captures the essence of Bellman's \emph{principle of
	optimality} \citep{bellman1954theory} i.e. we can solve a problem
by decomposing it into smaller\emph{ subproblems}, and the solutions
for these subproblems are combined to solve the original, larger problem.
In this context, constructing all combinations $\mathcal{S}^{n+m}$
for list $\mathcal{D}_{1}\cup\mathcal{D}_{2}$ is the problem; all
possible combinations $\mathcal{S}^{n}$, $\mathcal{S}^{m}$ for lists
$\mathcal{D}_{1}$, $\mathcal{D}_{2}$ respectively, are the smaller
subproblems.

In previous work of \citep{little2024polymor}, it was demonstrated
that the \emph{constraint algebra} (typically a \emph{group} or \emph{monoid})
can be implicitly embedded within a generator semiring through the
use of a \emph{convolution algebra}. Subsequently, the filter that
incorporates these constraints can be integrated into the generator
via the\emph{ semiring fusion theorem} (for details of semiring lifting
and semiring fusion, see \citep{little2024polymor}). In our case,
we can substitute the convolution algebra used in \citep{little2024polymor}
with the \emph{generator semiring} $\left(\left[\left[\mathbb{T}\right]\right],\cup,\circ,\emptyset,\left[\left[\;\right]\right]\right)$
($\mathbb{T}$ stands for type variable), then we have following equality

\begin{equation}
	\mathcal{S}_{k}^{n+m}=\underset{\begin{subarray}{c}
			i+j=k\\
			0\le i,j\le k
	\end{subarray}}{\bigcup}\left(\mathcal{S}_{i}^{n}\circ\mathcal{S}_{j}^{m}\right),\forall\;0\leq k\leq n+m\label{eq:sublists_merge_operation}
\end{equation}
where $l_{1}\circ l_{2}=\left[s\cup s^{\prime}\mid s\in l_{1},s^{\prime}\in l_{2}\right]$
is the \emph{cross-join operator} on lists $l_{1}$ and $l_{2}$ obtained
by concatenating each element $s$ in configuration $l_{1}$ with
each element $s^{\prime}$ in $l_{2}$. For instance, $\left[\left[1\right],\left[2\right]\right]\circ\left[\left[3\right],\left[4\right]\right]=\left[\left[1,3\right],\left[1,4\right],\left[2,3\right],\left[2,4\right]\right]$.
Informally, (\ref{eq:sublists_merge_operation}) is true because the
size $k$ combinations for the list $\mathcal{D}_{1}\cup\mathcal{D}_{2}$
should be constructed from all possible combinations $\mathcal{S}_{i}^{n}$
in $\mathcal{D}_{1}$ and $\mathcal{S}_{j}^{m}$ in $\mathcal{D}_{2}$
such that $i+j=k$ for all $0\leq i,j\leq k$. In other words, all
possible size $k$ combinations should be constructed by joining all
possible combinations with a size smaller than $k$.

Definition (\ref{eq:sublists_merge_operation}) is a special kind
of \emph{convolution} \emph{product} for two lists $l_{a}=\left[a_{0},a_{2},\ldots,a_{n}\right]$
and $l_{b}=\left[b_{0},b_{2},\ldots,b_{m}\right]$,
\begin{equation}
	\mathit{conv}\left(f,l_{a},l_{b},k\right)=\left[c_{0},c_{1},\ldots,c_{k}\right],
\end{equation}
where $c_{k}$ is defined as
\begin{equation}
	c_{k}=\underset{\begin{subarray}{c}
			i+j=k\\
			0\le i,j\le k
	\end{subarray}}{\bigcup}f\left(a_{i},b_{j}\right),\quad0\leq k\leq n+m.
\end{equation}

Given $\mathcal{S}^{n}$ and $\mathcal{S}^{m}$, it is not difficult
to verify that $\mathcal{S}^{n+m}=\mathit{conv}\left(\circ,\mathcal{S}^{n},\mathcal{S}^{m},n+m\right)$,
representing all sublists for list $\mathcal{D}_{1}\cup\mathcal{D}_{2}$.
Furthermore, all combinations with size smaller than $k$ can be obtained
by calculating $\mathit{conv}\left(\circ,\mathcal{S}^{n},\mathcal{S}^{m},k\right)$,
denoted as $\mathcal{S}_{\leq k}^{n+m}$. Using equation $\mathcal{S}_{\leq k}^{n+m}=\mathit{conv}\left(\circ,\mathcal{S}_{i}^{n},\mathcal{S}_{j}^{ m},k\right)$,
a combination generator recursion $\mathit{gen}_{\mathit{combs}}:\mathcal{N}\times\mathcal{K}\to\left[\mathbb{R}^{D}\right]$
can be defined as
\begin{equation}
	\begin{aligned}\mathit{gen}_{\mathit{combs}} & \left(0,k\right)=\mathit{merge}\left(\left[\;\right],k\right)\\
		\mathit{gen}_{\mathit{combs}} & \left(n,k\right)=\mathit{merge}\left(\mathit{merge}\left(\left[\boldsymbol{x}_{n}\right],k\right),\mathit{gen}_{\mathit{combs}}\left(n-1,k\right),k\right),
	\end{aligned}
	\label{eq:k-sublist-generator-parallel}
\end{equation}
which generate all combinations with a size $k$ or less. The $merge$
function is defined by \emph{pattern matching}\footnote{The definition of the function depends on the \emph{pattern}s (or
	\emph{cases)} of the input.}
\begin{equation}
	\begin{aligned}\mathit{merge} & \left(\left[\;\right],k\right)=\left[\left[\left[\;\right]\right]\right]\\
		\mathit{merge} & \left(\left[\boldsymbol{x}_{n}\right],k\right)=\left[\left[\left[\;\right]\right],\left[\left[\boldsymbol{x}_{n}\right]\right]\right]\\
		\mathit{merge} & \left(l,m,k\right)=\mathit{conv}\left(\circ,l,m,k\right)
	\end{aligned}
	.\label{eq:merge_operation}
\end{equation}

We denote the result of $\mathit{gen}_{\mathit{combs}}\left(n,k\right)$
for data sequence $\mathcal{D}_{i}$ by $\mathcal{S}_{\leq k}^{n}\left(\mathcal{D}_{i}\right)$.
For instance, $\mathit{gen}_{\mathit{combs}}\left(3,2\right)=\mathcal{S}_{\leq2}^{3}\left(\left[\boldsymbol{x}_{1},\boldsymbol{x}_{2},\boldsymbol{x}_{3}\right]\right)=\left[\left[\left[\right]\right],\left[\left[\boldsymbol{x}_{1}\right],\left[\boldsymbol{x}_{2}\right],\left[\boldsymbol{x}_{3}\right]\right],\left[\left[\boldsymbol{x}_{1},\boldsymbol{x}_{2}\right],\left[\boldsymbol{x}_{1},\boldsymbol{x}_{3}\right],\left[\boldsymbol{x}_{2},\boldsymbol{x}_{3}\right]\right]\right]$.
See figure \ref{fig:generic-recursive-generator } for an illustration
of the process by which recursive combination generator (\ref{eq:k-sublist-generator-parallel})
operates.

\subsection{Applying shortcut fusion equational transformations to derive an
	efficient, correct implementation\label{subsec:fusion}}

The combination generator $\mathit{gen}_{\mathit{combs}}$ constructed
above, gives us an efficient recursive basis to use \ref{eq:generate-and-test_paradigm}
to derive an efficient algorithm for solving the $K$-medoids problem.
Thus, a provably correct algorithm for solving the $K$-medoids problem
can be rendered as
\begin{equation}
	s^{*}=\mathit{sel}_{E}\left(\mathit{eval}_{E}\left(\mathit{gen}_{\mathit{combs}}\left(n,k\right)\right)\right).\label{eq:k-medoids_exhastive}
\end{equation}

As we mentioned above, if both selector $\mathit{sel}_{E}$ and evaluator
$\mathit{eval}_{E}$ can be fused into the generator by identifying
specific shortcut fusion theorems, then a significant amount of computational
effort and memory can be saved. Fortunately, both $\mathit{\mathit{sel}_{E}}$
and $\mathit{eval}_{E}$ are fusable in this problem. The aim is to
integrate all the functions in algorithm (\ref{eq:k-medoids_exhastive})
into one single, recursive function that will compute a globally optimal
solution $s^{*}$ for the $K$-medoids problem, efficiently.

The evaluator $\mathit{eval}$ is fusable with $\mathit{gen}$ because
the conditions for so-called \emph{tupling fusion }\citep{he2023efficient,little2024polymor}
hold. This form of fusion is applicable for many machine learning
problems because objective functions such as $E$ split up into a
\emph{sum of loss terms}, one per data item, so that it can be accumulated
during the generator recursion which visits each input item in the
input data $\mathcal{D}$ in sequence \citep{little2019machine}.
Using this, we fuse$\mathit{eval}_{E}$ inside the combination generator
$\mathit{gen}_{\mathit{combs}}$ by defining

\begin{equation}
	\begin{aligned}\mathit{evalgen}_{E,\mathit{combs}} & \left(0,k\right)=\mathit{merge}_{E}\left(\left[\;\right],k\right)\\
		\mathit{evalgen}_{E,\mathit{combs}} & \left(n,k\right)=\mathit{merge}_{E}\left(\mathit{merge}_{E}\left(\left[\boldsymbol{x}_{n}\right],k\right),\mathit{evalgen}_{E,\mathit{combs}}\left(n-1,k\right),k\right),
	\end{aligned}
	\label{eq:k-medoids_alg}
\end{equation}
where $\mathit{merge}_{E}$ is defined as

\begin{equation}
	\begin{aligned}\mathit{merge}_{E} & \left(\left[\;\right],k\right)=\left[\left[\left(\left[\;\right],\infty\right)\right]\right]\\
		\mathit{merge}_{E} & \left(\left[x_{n}\right],k\right)=\left[\left[\left(\left[\;\right],\infty\right)\right],\left[\left(\left[\boldsymbol{x}_{n}\right],\infty\right)\right]\right]\\
		\mathit{merge}_{E} & \left(l,m,k\right)=\mathit{conv}\left(\circ_{E},l,m,k\right),
	\end{aligned}
	\label{eq:eval_merge_operation}
\end{equation}
and $\circ_{E}$ is the cross-join operator described above augmented
with evaluation updates,
\begin{equation}
	l_{1}\circ_{E}l_{2}=\begin{cases}
		\left[\left(s_{1}\cup s_{2},E\left(s_{1}\cup s_{2}\right)\right)\mid\left(s_{1},r\right)\in l_{1},\left(s_{2},r\right)\in l_{2}\right] & \text{if \ensuremath{\left|s_{1}\cup s_{2}\right|=k}}\\
		\left[\left(s_{1}\cup s_{2},r\right)\mid\left(s_{1},r\right)\in l_{1},\left(s_{2},r\right)\in l_{2}\right] & \text{otherwise},
	\end{cases}
\end{equation}
which evaluates the objective value for configuration $s_{1}\cup s_{2}$
while concatenating the two corresponding configurations $\left(s_{1},r\right)$
and $\left(s_{2},r\right)$ such that $\left|s_{1}\cup s_{2}\right|=k$.
Thus we have achieved the fusion,
\begin{equation}
	\begin{aligned}s^{*} & =\mathit{sel}_{E}\left(\mathit{eval}_{E}\left(\mathit{gen}_{\mathit{combs}}\left(n,k\right)\right)\right)\\
		& =\mathit{sel}_{E}\left(\mathit{evalgen}_{E,\mathit{combs}}\left(n,k\right)\right).
	\end{aligned}
\end{equation}

Evaluating the complete configuration objective directly while generating
allows us to store only the currently best encountered configuration
on each recursive step. Finally, this allows us to fuse the selector
into the evaluator-generator $\mathit{evalgen}_{E,\mathit{combs}}$,
thus, only partial configurations with sizes smaller than $K-1$ are
stored, leading to lower space complexity of $O\left(N^{K-1}\right)$.
This strategy is well-suited to vectorized implementations and thus
also appropriate for parallel implementation.

Therefore, we use the following equational reasoning to derive our
EKM algorithm for solving the $K$-medoids problem

\begin{equation}
	\begin{aligned}s^{*} & =\mathit{sel}_{E}\left(\mathit{evalgen}_{E,\mathit{combs}}\left(n,k\right)\right)\\
		& =\mathit{selevalgen}_{E,\mathit{combs}}\left(n,k\right)\\
		& =\mathit{EKM}\left(n,k\right).
	\end{aligned}
	\label{eq:EKM}
\end{equation}

In summary, for clustering a length $N$ data input $\mathcal{D}$
into $K$ clusters, our derived algorithm $\mathit{EKM}\left(N,K\right)$
has worst-case $O\left(N^{K+1}\right)$ time complexity because recursion
$\mathit{selevalgen}_{E,\mathit{combs}}$ takes $N$ recursive steps
where in each step at most $O\left(N^{K}\right)$ configurations are
generated, thus the total time complexity of the algorithm is $O\left(N^{K+1}\right)$.
It is guaranteed to compute a globally optimal solution to (\ref{eq:K-medoids_MIP})
because it was derived from the exhaustive search algorithm (\ref{eq:generate-and-test_paradigm})
through correct, equational shortcut fusion transformations.

\section{Experiments}

In this section, we analyze the computational performance of our novel
exact $K$-medoids algorithm (EKM) on both synthetic and real-world
data sets. Our evaluation aims to test the following predictions:
(a) the EKM algorithm always obtains the best objective value\footnote{The squares Euclidean distance function was chosen for the experiments,
	any other proper metrics could also be used.}; (b) wall-clock runtime matches the worst-case time complexity analysis;
(c) We anlysis the wall-clock runtime comparing of EKM algorithm against
modern off-the-shell MIP solver (GLPK). The MIP solver present an
exponential complexity in the worst-case. 

We ran all our experiments by using the sequential version of our
algorithm, the numerous matrix operations required at every recursive
step are batch processed on the GPU. More sophisticated parallel programming
can be developed in the future. We executed all the experiments on
a Intel Core i9 CPU, with 24 cores, 2.4-6 GHz, and 32 GB RAM, and
GeForce RTX 4060 Ti GPU. 

\subsection{Real-world data set performance}

We test the perfermance of our EKM algorithm agianst various approximate
algorithms: Partition around medoids (PAM), Fast-PAM, Clustering Large
Applications based on RANdomized Search (CLARANS) on 18 datasets in
UCI Machine Learning Repository, two open-source datasets from \citep{wang2022predicting,padberg1991branch,ren2022global},
and 2 synthetic datasets. We show that our exact algorithm can always
deliver the best objective values among all other algorithms (see
Table \ref{tab:Empirical-comparison-on}).

To compare our algorithm's performance against the BnB algorithm proposed
by \citep{ren2022global}, we analyzed most of the real-world datasets
used in their study. We discovered that all these datasets (except
IRIS) can be solved exactly using either the PAM or Fast-PAM algorithms,
this indeed provids a extremly tight upper-bound in the analysis of
\citep{ren2022global}. Additionally, our experiments included real-world
datasets with a maximum size of $N=5000$. To best of our knowledge,
the largest dataset for which an exact solution has been previously
obtained is $N=150$, as documented by \citep{ceselli2005branch}.
Existing literatures on the $K$-medoids problem have only reported
exact solutions on small datasets, primarily due to the use of BnB
algorithms. Given their unpredictable runtime and exponential complexity
in worst-case scenarios, most BnB algorithms impose a time constraint
to avoid indefinite running times or memory overflow.

\begin{table}
    \scriptsize
		\caption{Empirical comparison of our novel exact $K$-medoids algorithm, EKM,
		against widely-used approximate algorithms (PAM, Fast-PAM, and CLARANS)
		for $K=3$, in terms of sum-of-squared errors ($E$), smaller is better.
		Best performing algorithm marked \textbf{bold}. Wall clock execution
		time in brackets (seconds). Datasets are from the UCI repository.
		\label{tab:Empirical-comparison-on}}
		
	\begin{centering}
	    \begin{tabular}{@{} >{\raggedright}p{0.04\textwidth}  >{\raggedleft}p{0.04\textwidth} >{\raggedleft}p{0.04\textwidth} >{\raggedleft}p{0.14\textwidth}>{\raggedleft}p{0.14\textwidth}>{\raggedleft}p{0.14\textwidth}>{\raggedleft}p{0.14\textwidth}@{}}
			\toprule
			UCI dataset & $N$ & $D$ & EKM (ours) & PAM & Faster-PAM & CLARANS\tabularnewline
         	\midrule
			LM  & 338 & 3 & \textbf{$\boldsymbol{3.96\times10^{1}}$} 
			
			($6.82$) & $3.99\times10^{1}$
			
			($4.02\times10^{-3}$) & $4.07\times10^{1}$
			
			($3.01\times10^{-3}$) & $5.33\times10^{1}$
			
			($6.14$)\tabularnewline
			\hline 
			UKM  & 403 & 5 & \textbf{$\boldsymbol{8.36\times10^{1}}$}
			
			($1.21\times10^{1}$) & $8.44\times10^{1}$
			
			($8.57\times10^{-3}$) & $8.40\times10^{1}$
			
			($3.21\times10^{-3}$) & $1.16\times10^{2}$ 
			
			($4.98\times10^{1}$)\tabularnewline
			\hline 
			LD  & 345 & 5 & $\boldsymbol{3.31\times10^{5}}$
			
			($6.98$) & $3.56\times10^{5}$
			
			($4.11\times10^{-3}$) & \textbf{$\boldsymbol{3.31\times10^{5}}$}
			
			($3.87\times10^{-3}$) & $4.68\times10^{5}$
			
			($3.40$)\tabularnewline
			\hline 
			Energy  & 768 & 8 & \textbf{$\boldsymbol{\boldsymbol{2.20}\times10^{6}}$}
			
			($1.01\times10^{2}$) & $2.28\times10^{6}$
			
			($6.95\times10^{-3}$) & $2.28\times10^{6}$
			
			($3.94\times10^{-3}$) & $2.97\times10^{6}$
			
			($2.71$) \tabularnewline
			\hline 
			VC  & 310 & 6 & \textbf{$\boldsymbol{3.13\times10^{5}}$}
			
			($4.98$) & \textbf{$\boldsymbol{3.13\times10^{5}}$}
			
			($3.15\times10^{-3}$) & $3.58\times10^{5}$
			
			($5.36\times10^{-3}$) & $5.27\times10^{5}$
			
			($2.58$)\tabularnewline
			\hline 
			Wine  & 178 & 13 & \textbf{$\boldsymbol{2.39\times10^{6}}$} 
			
			($1.11$) & \textbf{$\boldsymbol{2.39\times10^{6}}$}
			
			($1.06\times10^{-3}$) & $2.63\times10^{6}$
			
			($2.34\times10^{-3}$) & $6.86\times10^{6}$
			
			($5.56\times10^{-1}$)\tabularnewline
			\hline 
			Yeast & 1484 & 8 & \textbf{$\boldsymbol{8.37\times10^{1}}$}
			
			($1.10\times10^{3}$) & $8.42\times10^{1}$
			
			($9.54\times10^{-2}$) & \textbf{$8.42\times10^{1}$}
			
			($6.08\times10^{-2}$) & \textbf{$1.05\times10^{2}$}
			
			($1.73\times10^{2}$)\tabularnewline
			\hline 
			IC & 3150 & 13 & \textbf{$\boldsymbol{6.9063\times10^{9}}$}
			
			($2.46\times10^{4}$) & \textbf{$6.9105\times10^{9}$}
			
			($8.68\times10^{-1}$) & \textbf{$\boldsymbol{6.9063\times10^{9}}$}
			
			($1.91\times10^{-1}$) & \textbf{$1.44\times10^{10}$}
			
			($2.70\times10^{1}$)\tabularnewline
			\hline 
			WDG & 5000 & 21 & \textbf{$\boldsymbol{1.67\times10^{5}}$}
			
			($2.49\times10^{5}$) & \textbf{$\boldsymbol{1.67\times10^{5}}$}
			
			($1.34$) & \textbf{$\boldsymbol{1.67\times10^{5}}$}
			
			($1.97\times10^{-1}$) & \textbf{$2.77\times10^{5}$}
			
			($5.32\times10^{3}$)\tabularnewline
			\hline 
			IRIS & 150 & 4 & $\boldsymbol{8.40\times10^{1}}$
			
			($7.32\times10^{-1}$) & $8.45\times10^{1}$
			
			($2.51\times10^{-3}$) & $8.45\times10^{1}$
			
			($1.03\times10^{-3}$) & $1.57\times10^{2}$ 
			
			($2.32\times10^{-1}$)\tabularnewline
			\hline 
			SEEDS & 210 & 7 & \textbf{$\boldsymbol{5.98\times10^{2}}$}
			
			($1.71$) & \textbf{$\boldsymbol{5.98\times10^{2}}$}
			
			($1.14\times10^{-3}$) & \textbf{$\boldsymbol{5.98\times10^{2}}$ }
			
			($3.59\times10^{-3}$) & $1.12\times10^{3}$ 
			
			($7.82\times10^{-1}$) \tabularnewline
			\hline 
			GLASS & 214 & 9 & \textbf{$\boldsymbol{6.29\times10^{2}}$}
			
			($1.81$) & \textbf{$\boldsymbol{6.29\times10^{2}}$}
			
			($1.01\times10^{-3}$) & \textbf{$\boldsymbol{6.29\times10^{2}}$ }
			
			($1.62\times10^{-3}$) & $1.04\times10^{3}$
			
			($2.27$)\tabularnewline
			\hline 
			BM & 249 & 6 & \textbf{$\boldsymbol{8.63\times10^{5}}$}
			
			($2.64$) & $8.76\times10^{5}$
			
			($4.12\times10^{-3}$) & \textbf{$\boldsymbol{8.63\times10^{5}}$}
			
			($1.61\times10^{-3}$) & $1.33\times10^{6}$
			
			($1.02\times10^{1}$)\tabularnewline
			\hline 
			HF & 299 & 12 & \textbf{$\boldsymbol{7.83\times10^{11}}$}
			
			($4.57$) & \textbf{$\boldsymbol{7.83\times10^{11}}$}
			
			($1.00\times10^{-3}$) & \textbf{$\boldsymbol{7.83\times10^{11}}$}
			
			($4.87\times10^{-3}$) & \textbf{$1.88\times10^{12}$}
			
			($6.55\times10^{-1}$) \tabularnewline
			\hline 
			WHO & 440 & 7 & $\boldsymbol{8.33\times10^{10}}$
			
			($1.67\times10^{1}$) & $\boldsymbol{8.33\times10^{10}}$
			
			($5.62\times10^{-3}$) & $\boldsymbol{8.33\times10^{10}}$
			
			($2.81\times10^{-3}$) & $1.21\times10^{11}$
			
			($8.12$)\tabularnewline
			\hline 
			ABS & 740 & 19 & \textbf{$\boldsymbol{2.32\times10^{6}}$}
			
			($1.04\times10^{2}$) & \textbf{$\boldsymbol{2.32\times10^{6}}$}
			
			($2.11\times10^{-2}$) & \textbf{$2.38\times10^{6}$}
			
			($5.00\times10^{-3}$) & $2.96\times10^{6}$
			
			($7.80\times10^{1}$)\tabularnewline
			\hline 
			TR & 980 & 10 & \textbf{$\boldsymbol{1.13\times10^{3}}$}
			
			($2.16\times10^{2}$) & \textbf{$\boldsymbol{1.13\times10^{3}}$}
			
			($5.14\times10^{-2}$) & \textbf{$\boldsymbol{1.13\times10^{3}}$}
			
			($1.14\times10^{-2}$) & \textbf{$1.38\times10^{3}$}
			
			($2.59\times10^{2}$)\tabularnewline
			\hline 
			SGC & 1000 & 21 & $\boldsymbol{1.28\times10^{9}}$
			
			($2.20\times10^{2}$) & $\boldsymbol{1.28\times10^{9}}$
			
			($1.71\times10^{-1}$) & $\boldsymbol{1.28\times10^{9}}$
			
			($4.22\times10^{-2}$) & $2.52\times10^{9}$
			
			($2.24$)\tabularnewline
			\hline 
			HEMI & 1995 & 7 & $\boldsymbol{9.91\times10^{6}}$
			
			($3.06\times10^{3}$) & $\boldsymbol{9.91\times10^{6}}$
			
			($3.64\times10^{-1}$) & $\boldsymbol{9.91\times10^{6}}$
			
			($6.99\times10^{-2}$) & $1.66\times10^{7}$
			
			($9.53$)\tabularnewline
			\hline 
			PR2392 & 2392 & 2 & \textbf{$\boldsymbol{2.13\times10^{10}}$}
			
			($1.38\times10^{4}$) & \textbf{$\boldsymbol{2.13\times10^{10}}$}
			
			($3.66\times10^{-1}$) & \textbf{$\boldsymbol{2.13\times10^{10}}$}
			
			($8.38\times10^{-2}$) & \textbf{$3.47\times10^{10}$}
			
			($1.15\times10^{2}$)\tabularnewline
	        \bottomrule
		\end{tabular}
		\par\end{centering}

\end{table}

\subsection{Time complexity analysis for serial implementation}

\begin{figure}
	\begin{centering}
		\includegraphics[scale=0.35]{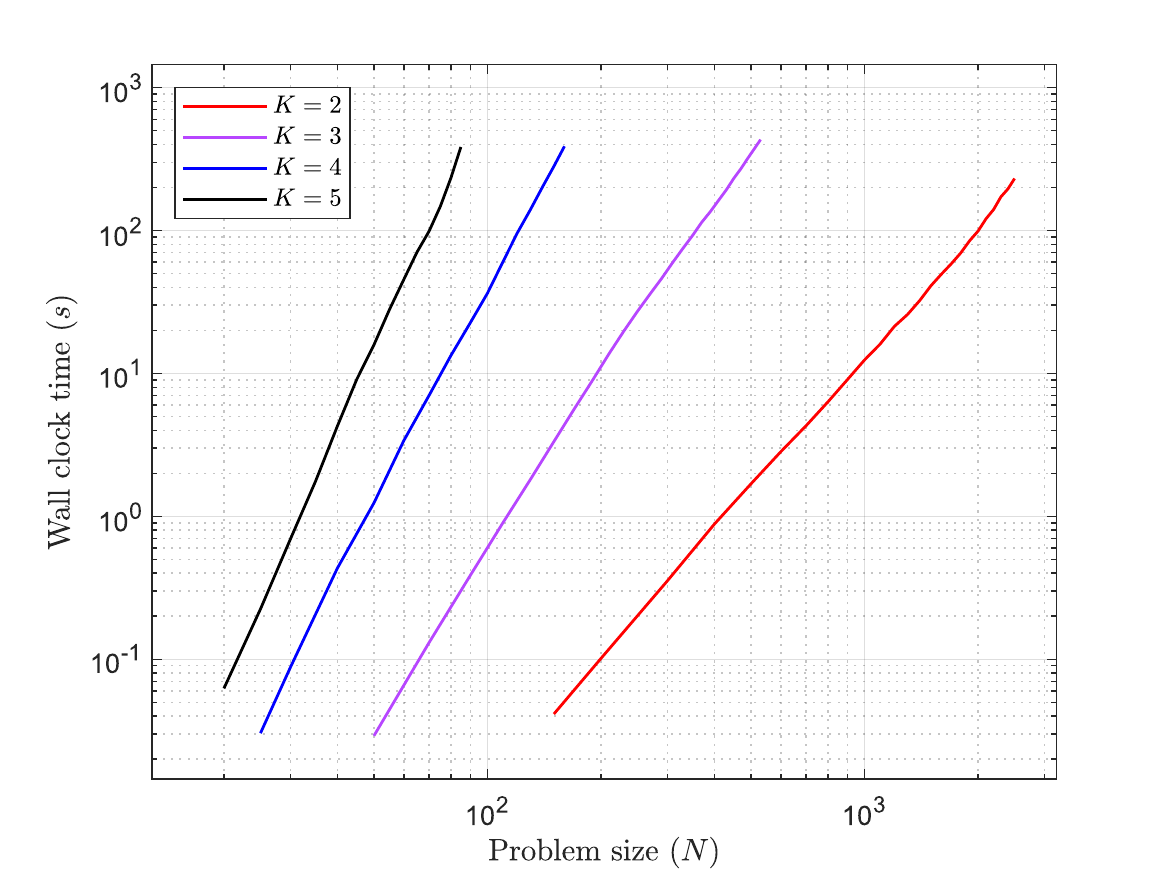}\includegraphics[scale=0.35]{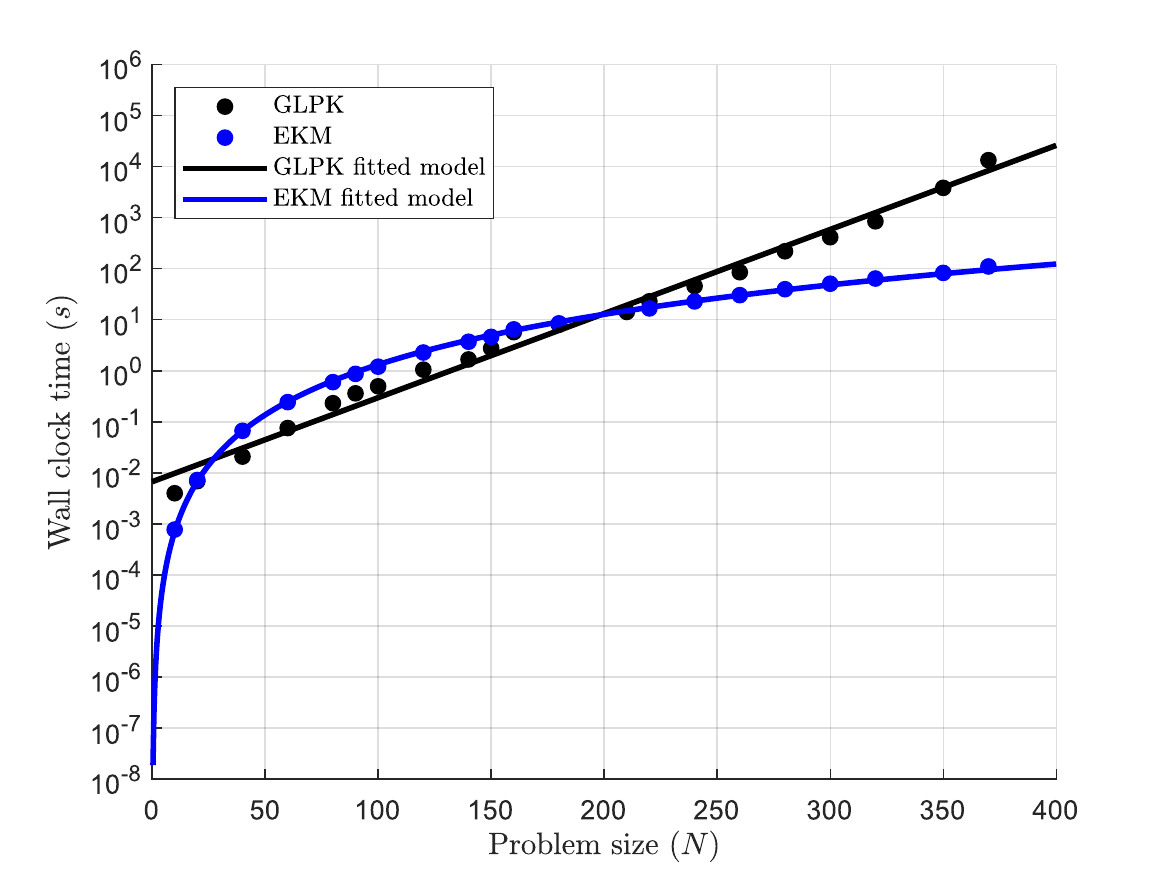}
		\par\end{centering}
	\caption{Log-log wall-clock run time (seconds) for our algorithm (EKM) tested
		on synthetic datasets (left panel). The run-time curves from left
		to right (corresponding to $K=2,3,4,5$ respectively), have slopes
		3.005, 4.006, 5.018, and 5.995, an excellent match to the predicted
		worst-case run-time complexity of $O\left(N^{3}\right)$, $O\left(N^{4}\right)$,
		$O\left(N^{5}\right)$, and $O\left(N^{6}\right)$ respectively. Log-linear
		wall-clock run-time (seconds) comparing EKM algorithm against a classical
		MIP (BnB) solver (GLPK) on synthetic datasets with $K=3$ (right panel).
		On this log-linear scale, exponential run-time appears as a linear
		function of problem size $N$, whereas polynomial run-time is a logarithmic
		function of $N$. \label{fig:log-log-wall-clock-run}}
\end{figure}

We test the wall-clock time of our novel EKM algorithm on a synthetic
dataset with cluster sizes ranging from $K=2$ to $5$. When $K=2$,
the data size $N$ ranges from $150$ to 2,500, $K=3$ ranges from
50 to 530, $K=4$ ranges from 25 to 160, and $K=5$ ranges from 30
to 200. The worst-case predictions are well-matched empirically (figure
\ref{fig:log-log-wall-clock-run}, left panel). As predicted, the
off-the-shelf BnB-based MIP solver (GLPK) exhibits worst-case exponential
time complexity (figure \ref{fig:log-log-wall-clock-run}, right panel).

\section{Discussion}

As our predictions and experiments show, our novel EKM algorithm clearly
outperforms all other algorithms which can be guaranteed to obtain
the exact globally optimal solution to the $K$-medoids problem. We
also demonstrated for the first time that for many of the datasets
used as test cases for this problem, approximate algorithms can apparently
obtain exact solutions. It would not be possible to know this without
the computational tractability of EKM.

Besides presenting our novel EKM algorithm, we aim to prompt researchers
to reconsider the metric for assessing the efficiency of exact algorithms
to account for subtleties beyond simple problem scale. Although many
ML problems may be NP-hard in their most general form, they are often
not specified in full generality. In such cases (which are quite common
in practice), polynomial-time solutions may be available. Therefore,
comparing algorithms solely on the basis of problem scale is neither
fair nor accurate, as it ignores variations in actual run-times, such
as the tightness of upper bounds, or parameters independent of problem
scale. Additionally, as the scale of problems increases, the distinction
between exact and approximate algorithms tends to diminish. We observe
that this is indeed a common phenomenon in the study of exact algorithms.
Therefore, when dealing with sufficiently large datasets, the use
of exact algorithms may be unnecessary. We discuss this topic in more
detail next.

In discussing the ``\emph{goodness}'' of exact algorithms for ML,
it is critical to recognize that focusing solely on the scalability
of these algorithms---for instance, their capacity to handle large
datasets---does not provide a comprehensive assessment of their utility.
This inclination to prioritize scalability when assessing exact algorithms
arises from the perceived intractable combinatorics of many ML problems,
and most of these problems are classified as NP-hard, so that no known
algorithm can solve all instances of the problem in polynomial time.

However, for many ML problems, the problems specified for proving
NP-hardness are not the same as their original definitions used in
practical ML applications. Apart from the $K$-medoids problem, polynomial-time
algorithms for solving the 0-1 loss classification problem \citep{he2023efficient}
and other $K$-clustering problems \citep{inaba1994applications,tirnuaucua2018global}
have also been developed in the literature. If a polynomial-time algorithm
does exist for these seemingly intractable problems, overemphasizing
scalability can mislead scientific development, diverting attention
from important measures such as memory usage, worst-case time complexity,
and the practical applicability of the algorithm in real-world scenarios.
For example, by setting the cluster size to one, the $K$-medoids
problem can be solved exactly by choosing the closest data item to
the mean of the data set, a strategy with $O\left(N\right)$ time
complexity in the worst-case. While this $O\left(N\right)$ time algorithm
can efficiently handle very large scale datasets, it does little to
advance our understanding of the fundamental principles involved.

Therefore, judging an algorithm implementation solely by the scale
of the dataset it can process is not an adequate measure of its effectiveness.
Indeed, for large datasets, the use of exact algorithms may often
be unnecessary as many high-quality approximate algorithms provide
very good results, supported by solid theoretical assurances. If the
clustering model closely aligns with the ground truth, the discrepancy
between approximate and exact solutions should not be significant,
provided the dataset is sufficiently large. Thus, it is not surprising
that \citep{ren2022global}'s algorithm can achieve excellent approximate
solutions with more than one million data points, a typical occurrence
in studies involving exact algorithms. Past research has shown that
while exact algorithms can quickly find solutions, most of the effort
is expended on verifying their optimality \citep{dunn2018optimal,ustun2017simple}.
It is possible that the first configuration generated by the algorithm
is optimal, but proving its optimality without exploring the entire
solution space $\mathcal{S}$ is impossible (unless additional information
is provided).

For the study of the $K$-medoids problem, while algorithms presented
in \citep{ren2022global,ceselli2005branch,elloumi2010tighter,christofides1982tree}
are exact in principle, experiments reported by the authors do not
demonstrate the actual computation of exact solutions, nor do they
provide any theoretical guarantee on the computational time required
to achieve satisfactory approximate solutions. If the application
of the problem is concerned with only the approximate solution, it
may be more beneficial to concentrate on developing more efficient
or more robust heuristic algorithms. This could potentially offer
more practical value in scenarios where approximate solutions are
adequate.

\section{Conclusions and future work}

In this paper we derived EKM, a novel exact algorithm for the $K$-medoids
clustering problem with worst-case $O\left(N^{K+1}\right)$ time complexity.
EKM is also easily parallelizable. EKM is the first rigorously proven
approach to solve the $K$-medoids problem with both polynomial time
and space complexities in the worst-case. This enables precise prediction
of time and space requirements before attempting to solve a clustering
problem, in contrast to existing BnB algorithms which in practice
require a hard computation time limit to prevent memory overflow or
bypass the exponential worst-case run time. To demonstrate the effectiveness
of this algorithm, we applied it to various real-world datasets and
achieved exact solutions for datasets significantly larger than those
previously reported in the literature. In our experiments we were
able to process datasets of up to $N=5,000$ data items, a considerable
increase from the previous maximum of around $N=150$. The main disadvantage
of our algorithm is that its space and time complexity is polynomial
with respect to the number of medoids, $K$. Thus for problems that
involve a large number of medoids, our algorithm quickly becomes intractable.
Currently, we run our algorithm sequentially in experiments, with
the very large number of matrix operations processed by GPU. This
approach is far from the ideal parallel implementation. In the future,
a more sophisticated parallelizable implementation could be developed;
the inherently parallel structure of our combinations generator makes
this a relatively simple prospect.

With exact solutions for combinatorial ML problems, the memory-computation
trade-off is always present. However, with BnB algorithms and MIP
solvers, space complexity analysis is often omitted making it difficult
to ascertain the actual memory requirements. Specifically, when using
off-the-shelf MIP solvers like GLPK or Gurobi, the memory required
just to specify the problem can be substantial. For instance, to describe
the $K$-medoids problem in MIP form (\ref{eq:K-medoids_MIP}) a constraint
matrix\footnote{There are $N^{2}$ constraints to ensure each point is assigned to
	exactly one cluster, $N$ constraints to identify which data items
	are medoids, and one additional constraint for the number of medoids} of size $\left(N^{2}+N+1\right)\times N$ is required \citep{ren2022global,vinod1969integer}.
Indeed, memory overflow issues have been reported in almost all practical
usage of BnB algorithms \citep{ceselli2005branch,elloumi2010tighter,christofides1982tree}.
Therefore, setting a computation time limit is a necessary restriction
on the use of BnB algorithms, a restriction which only applies to
EKM at large values of $K$.

One way to reduce memory usage in our algorithm is by representing
configurations in compact \emph{integer} or \emph{bit format}. For
instance, a sublist requires $O\left(N\right)$ bytes but its integer
representation can require a much smaller $O\left(\log N\right)$
bytes. Indeed, if the generated configurations are arranged in a special
order, then the transformation between the configuration and its integer
representation can be achieved in constant time using \emph{ranking}
and \emph{unranking} functions. For instance, the \emph{revolving-door
	ordering} is widely used for generating combinations \citep{ruskey2003combinatorial,kreher1999combinatorial}.

In real-world applications, problems are often characterized by additional
combinatorial constraints that must be satisfied as part of the MIP.
For example, we might want to limit the number of data items in any
cluster to a specified maximum or minimum. While the MIP specification
of this constraint is straightforward, it is difficult to efficiently
incorporate these constraints into a BnB algorithm in a predictable
way. However, in our framework, we can easily incorporate these types
of constraints into our generator using the semiring lifting technique
and the constrained problem can still be solved in polynomial time
in the worst-case. Details on applying semiring lifting techniques
are described in \citep{little2024polymor}.

Finally, when evaluating the objective value for each combination
we can incrementally update partial configurations, which requires
fewer operations because non-optimal configurations can be eliminated
early in the recursion by applying a \emph{global upper bound} which
can be obtained by any heuristic/approximate algorithm such as PAM
\citep{he2023efficient}. Past research has shown that this can achieve
up to an order of magnitude decrease in computational time required
to find a globally optimal solution.

\newpage

\bibliographystyle{plainnat}
\bibliography{Bibliography}

\end{document}